# A Probabilistic Network of Predicates

**Dekang Lin**
Department of Computer Science
University of Manitoba
Winnipeg, Manitoba, Canada, R3T 2N2
lindek@cs.umanitoba.ca

## Abstract

Bayesian networks are directed acyclic graphs
representing independence relationships among
a set of random variables. A random variable
can be regarded as a set of exhaustive and mutu-
ally exclusive propositions. We argue that there
are several drawbacks resulting from the propo-
sitional nature and acyclic structure of Bayesian
networks.

To remedy these shortcomings, we propose
a probabilistic network where nodes represent
unary predicates and which may contain directed
cycles. The proposed representation allows us
to represent domain knowledge in a single static
network even though we cannot determine the
instantiations of the predicates before hand. The
ability to deal with cycles also enables us to han-
dle cyclic causal tendencies and to recognize re-
cursive plans.

## 1 INTRODUCTION

Bayesian networks [Pearl, 1988] are directed acyclic graphs
which encode independence relationships among random
variables represented by the nodes. Applications of
Bayesian networks have been found in diagnosis [Pearl,
1988; Heckerman, 1990], decision analysis [Henrion et al.,
1991], story understanding [Charniak and Goldman, 1989;
Charniak and Goldman, 1991], etc.

Reasoning in Bayesian networks takes the form of assign-
ing values to a set of random variables in the network
and then propagating the influence of these assignments
to other nodes in the network. One of the important kinds
of inference performed by Bayesian networks is abduc-
tion. Abduction is the inference to the best explanation.
Many AI tasks such as diagnosis [Peng and Reggia, 1990;
Lin and Goebel, 1991a], plan recognition [Kautz, 1987;
Lin and Goebel, 1991b] can be formulated as abduction.
In Bayesian networks, observations are be represented by
assignments to a subset of variables in the network. An
explanation of the observations is an assignment of all the
variables in the network [Charniak and Goldman, 1991;

Pearl, 1988] or the subset of variables in the network that
are relevant to the observations [Shimony, 1991].

### 1.1 DRAWBACKS OF BAYESIAN NETWORK

Bayesian network formulation of abduction offers many ad-
vantages over other models of abduction. Abduction is un-
sound inference. There are typically multiple explanations
for any given set of observations. Bayesian network ap-
proach defines the best explanation to be the most probable
one. Goldman and Charniak argued [Charniak and Gold-
man, 1991] that the disadvantage of many non-probabilistic
approaches to plan recognition is that "it cannot decide that
a particular plan, no matter how likely, explains a set of
actions, as long as there is another plan, no matter how
unlikely, which also explain the actions."

Bayesian networks are acyclic graphs. Each random vari-
able represented by a node in a Bayesian network can be
regarded as a set of exhaustive and mutually exclusive
propositions. There are several drawbacks as a result of
the acyclicity and the propositional nature of Bayesian net-
works:

**Multiple instances**

In cases where there are multiple instances of the same type
of events, each event has to be represented by a separate
node in a Bayesian network. Analogous to the limitation of
propositional logic, propositional Bayesian network must
represent common properties of the same types of events
separately and explicitly for each instance. In medical di-
agnosis, there are typically at most one event of a particular
type in each case. However, consider this:

> Arnold dropped in at Bob's house. Bob was
> sneezing and having a headache. When Arnold
> went back home, he also began to sneeze and
> have headaches.

An explanation of this is that Bob had a flu and, since flu
is contagious, Arnold got it from Bob. The knowledge
used to make this inference include that flu causes sneez-
ing and headache and that flu is contagious. This knowl-
edge, however, is difficult to encode in a Bayesian net-
work, because we cannot determine the instances of flu and sneezing events before hand. Therefore, a Bayesian network must be dynamically constructed for each case [Charniak and Goldman, 1991; Goldman and Charniak, 1990; Horsch and Poole, 1990].

The disadvantage of dynamic construction is that the network construction mechanism has to depend on another representation scheme such as a deductive database [Charniak and Goldman, 1991; Goldman and Charniak, 1990] or logic schemas [Horsch and Poole, 1990] to represent domain knowledge. It then has to identify what is the relevant part of knowledge with regard to the current case. One of the most important advantages of Bayesian network over other representation schemes is that it is easier to identify what's relevant or not. Dynamic construction loses this advantage. Further more, the dynamically constructed Bayesian network is often, in effect, a representation of all the possible solutions to the problem. Bayesian network is only used to choose the most probable among the all possible ones. This reduce the usefulness of Bayesian network significantly.

### Acyclicity

Since Bayesian networks are acyclic graphs, they cannot represent recursive relations. In plan recognition, a plan may contain another plan of the same type as one of its components. For example, a plan to sort a list involves sorting sublists and then merge the results. One may argue that if the links in Bayesian network represent causal relationships, then the acyclic assumption does not affect the expressive power, because causal relationships are acyclic. We believe that this not necessarily true. There are two notions of causation: **token causation** and **causal tendency**. Token causation refers to a particular instance of causation, happening at a unique spatio-temporal location. For example, "Socrates's drinking of hemlock caused his death" is a token causation. Causal tendency, other hand, refers the tendency of events of one **type** to cause events of another **type**. For example, "flu tends to cause sneezing" is a causal tendency.

While token causations are acyclic, causal tendencies can be cyclic. For example, one person's flu may cause another person to catch flu. The cause and the effect are of the same type. The relationship between hens and eggs provides another example: hens engender eggs, and eggs hens.

### Imprecise Observation

The inputs to Bayesian network inference are variable assignments. Sometimes, however, an observation may not be in enough detail to warrant a variable assignment. For example: in auto-engine diagnosis, one of the most common symptoms is abnormal sound. There are several attributes that characterize different kinds of abnormal sound: the sound may be banging or pinging; its occurrence may be continual or intermittent. These different attribute values have different implications in diagnosis. A Bayesian network either represents abnormal sound as a single variable which takes values from the different combinations of the attributes, or represents each combinations of attributes as a single variable. In both cases, an observation of abnormal sound must specify every one of its attribute values, e.g., a continual, pinging sound. However, a novice may not be able to make these distinctions. He may only be able to tell the sound is abnormal. Such an observation is not in enough detail to assign a value to a variable in the network.

### Overly specific explanations

An explanation in Bayesian network is an assignment to the variables in the network. A problem with this definition of explanations is that explanations may be overly specific. For example, Figure 1 shows a Bayesian network. Supposing Mr. Holmes receives a message on his desk that his house alarm went off. The most likely explanation is that someone broke into his house and triggered the alarm. One of his concerned neighbour then phoned his office. Another explanation, which is less likely, is that the secretary played a practical joke on him and gave him this phony message. However, if Mr. Holmes has 50 neighbours, the node "neighbour's call" has 50 different values. Each break-in explanation must commit to one of them. As a result, each break-in explanation is less likely than the practical joke explanation even though break-in is more likely to be the case. This is obviously a situation we want to avoid. Shimony [Shimony, 1991] proposed assigning values to variables only to the relevant variables in the network. His proposal, however, will not resolve the problem in our example, because neighbour's call is relevant. One may wonder why should we distinguish between 50 different neighbours. The answer is that which person phoned may indeed be important. For example, if Mr. Holmes has a neighbour Jack who is notorious for making things up and Mr. Holmes knows the message is from Jack, then he might have more doubt on whether there was a break in than he would otherwise.

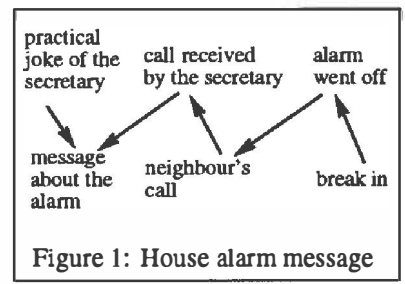

Figure 1: House alarm message

## 1.2 A PREDICATIVE APPROACH

In this paper, we introduce a probabilistic network for abductive reasoning, where nodes are interpreted as unary predicates instead of propositions. There is no acyclic restriction on the network. An observation is a description of an event that includes the type of the event, and a set of attribute values of the event. An explanation of the observations is a scenario which contains events that match the descriptions in the observations.

The domain knowledge is represented by a single static network. Given a set of observations, the inference algorithm retrieves the most probable explanation from the network. The explanation may contain multiple instantiations of the same node. Similar to Bayesian networks, the best explanation is defined to be the most probable one. We show that our approach overcomes the difficulties discussed earlier.

## 2    EVENT NETWORK

In our approach, the domain is partitioned into events and their attribute values. Attributes are characteristics that can be ascribed to events. Attributes are represented by functions that map events to attribute values. For example, **agent** is an attribute. If $x$ is an event, then $\text{agent}(x)$ is the agent of $x$.

An event type is a unary predicate over the set of events. For example, **flu** is an event type, and $\text{flu}(x)$ is true iff $x$ is a **flu** event. The event types are organized in generalization-specialization hierarchies.

The structural/causal relationships between events are represented by unary functions, called features. When the interpretation of a feature is causal, it represent the causal tendency between the two types of events that are the domain and the range of the feature function. For example, since **flu** tends to cause **sneezing**, there is a feature **sneeze-effect** that maps a **flu** event to a **sneezing** event. That is **sneeze-effect**$(x)$ is the **sneezing** effect of a **flu** event $x$.

There is a distinguished element in the domain to represent the intuitive notion of undefined. For example, if **flu** event $x$ did not cause a **sneezing** event, then **sneeze-effect**$(x) = \bot$.

The events, features and attributes are axiomatized in the language **Lp**, an extension of the first order logic (FOL) developed by Bacchus [Bacchus, 1988] to represent and reason with both logical and probabilistic information.

The taxonomic relationships between event types are represented by abstraction axioms, which are sentences in the form:

$$\forall x . c_1(x) \rightarrow c_2(x)$$

where $c_1$ and $c_2$ are event types. For example, that **flu** is a subclass of **disease** is represented by the sentence:

$$\forall x . \text{flu}(x) \rightarrow \text{disease}(x)$$

For each feature $f$ that maps events of type $c_1$ to events of type $c_2$, there is a feature restriction axiom:

$$\forall x . c_1(x) \wedge f(x) \neq \bot \rightarrow c_2(f(x))$$

The relationships between event types forms an event network.

**Definition 2.1 (Event Network)** *A event network is a directed graph whose nodes represent event types and whose links are as follows:*

**"isa" and specialization links:** *Corresponding to each abstraction axiom* $\forall x . c_1(x) \rightarrow c_2(x)$, *there is an "isa" link from $c_1$ to $c_2$ with the label "isa", and a specialization link from $c_2$ to $c_1$ with the label "spec."*

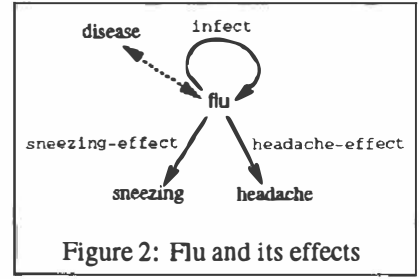

Figure 2: Flu and its effects

**feature links:** *Corresponding to each feature restriction axiom*

$$\forall x . (c_1(x) \wedge f(x) \neq \bot) \rightarrow c_2(f(x))$$

*there is a feature link from $c_1$ to $c_2$ and labeled $f$.*

Figure 2 shows an example. The solid links represent feature links. A bidirectional broken line represents a pair of "isa" and specialization links. The upward arrow represents the "isa" link and the downward arrow represents the specialization link.

A major source of complexity for many abductive reasoning systems is consistency checking. In our representation scheme, we restrict the consistency conditions to constraints on attribute values so that the consistency can be established efficiently.

There are two kinds of attribute value constraints: local constraints and percolation constraints.

Local constraints specify the relationships between the attributes of a single event and are expressed in the sentences in the following form:

$$\forall x . c(x) \rightarrow r(a_1(x), \ldots, a_k(x))$$

where $c$ is an event type and $a_1, \ldots, a_k$ are attributes, $r$ is a k-nary relation over attribute values. For example, that the **infectee** of a **flu** event is a different person than the **agent** can be represented by the local constraint:

$$\forall x . \text{flu}(x) \rightarrow \text{agent}(x) \neq \text{infectee}(x)$$

Percolation constraints specify the relationships between of the attributes of an event and its direct features. For example, suppose **infect** is the feature that represent **flu** causing **flu**, then we have the following percolation constraint:

$$\forall x, v . \text{flu}(x) \wedge \text{flu}(\text{infect}(x)) \wedge \text{agent}(\text{infect}(x)) = v$$
$$\rightarrow \text{infectee}(x) = v$$

Intuitively, this says that the agent of a **flu** event that is caused by another flu is the value of **infectee** attribute of the cause event. In general, percolation constraints are in the form:

$$\forall x, v . (c_1(x) \wedge c_2(f(x)) \wedge a'(f(x)) = v) \rightarrow a(x) = v$$

where $c_1, c_2$ are event types, $f$ is feature of $c_1$, $a, a'$ are attributes.

Together, the above examples of local and percolation constraints state that the agents of two flu events where one causing another are different.

So far, the axioms have all been FOL formulas. **Lp**'s extension to FOL is the ability to represent and reason with probabilistic knowledge, which are represented by probabilistic terms. Somewhat simplified, a probabilistic term in **Lp** is a formula in the form: $[\alpha(x)]_x$, where $\alpha(x)$ is a FOL formula with open variable $x$. It denotes the probability of a random individual in the domain satisfies the formula $\alpha(x)$. For example, $[\mathtt{flu}(x)]_x$ is the probability of an random event in the domain being a **flu** event. Conditional probabilities can also be conveniently expressed in **Lp**: $[\beta(x)|\alpha(x)]_x$ denotes the conditional probability of $\beta$ being true of a random element $x$ in the domain, given $\alpha(x)$ is true.

The probabilities we use to compute the probabilities of explanations are prior and conditional probabilities of event descriptions.

**Definition 2.2 (Event description)** *An event description specifies the type of an event and its attribute values. It is a formula in the following form:*

$$D(x) \equiv c(x) \wedge V(x)$$

*where c is a event type and $V(x)$ is a conjunction of attribute value assignments: $V(x) \equiv a_1(x) = v_1 \wedge a_2(x) = v_2 \wedge \ldots a_k(x) = v_k$, where $a_1, a_2, \ldots, a_k$ are attributes and $v_1, v_2, \ldots, v_k$ are attribute values. An event x is said to match a description D if $D(x)$ is true.*

The probabilistic knowledge in the domain are represented as category statistics in either of the following two forms:

$$[D(x)]_x = p \quad \text{or} \quad [D'(f(x))|D(x)]_x = p$$

where $D(x) = c(x) \wedge V(x)$ and $D'(x) = c'(x) \wedge V'(x)$ are event descriptions. $f$ is a feature that maps events in $c$ to $c'$.

In summary, the domain knowledge is represented by four groups of axioms: Abstraction axioms, feature restriction axioms, attribute value constraints, and category statistics.

## 3  SCENARIOS AS EXPLANATIONS

An observation is an existentially quantified event description. For example, the story about Arnold and Bob contains four observations:

$\exists x.\mathtt{sneezing}(x) \wedge \mathtt{agent}(x) = \mathtt{Bob} \wedge \mathtt{time}(x) = 1$
$\exists x.\mathtt{headache}(x) \wedge \mathtt{agent}(x) = \mathtt{Bob} \wedge \mathtt{time}(x) = 1$
$\exists x.\mathtt{sneezing}(x) \wedge \mathtt{agent}(x) = \mathtt{Arnold} \wedge \mathtt{time}(x) = 2$
$\exists x.\mathtt{headache}(x) \wedge \mathtt{agent}(x) = \mathtt{Arnold} \wedge \mathtt{time}(x) = 2$

An observation can be explained by hypothesizing that the event is a component or caused by another event. Such a relationship forms a directed path in the event network. The "isa" and specialization links in the event network allows one event to inherit causal/structural properties and explanations of its superclasses. Similar to inheritance reasoning in semantic networks, we use path preemption to ensure that information inherited is the most specific.

We denote the closure and the non-empty closure of "isa" links by $\xrightarrow{\text{isa}}*$ and $\xrightarrow{\text{isa}}+$, respectively. The meanings of $\xrightarrow{\text{spec}}*$ and $\xrightarrow{\text{spec}}+$ are similarly defined.

**Definition 3.1 (Path Preemption)**

**Generalization Preemption:** *A path $c \xrightarrow{\text{isa}}*c_2 \xrightarrow{f} c'$ is preempted if there exists a path $c \xrightarrow{\text{isa}}*c_1 \xrightarrow{f} c'_1 \xrightarrow{\text{isa}}*c'$ such that $c_1 \xrightarrow{\text{isa}}*c_2$ (Figure 3.a).*

**Specialization Preemption:** *A path $c \xrightarrow{f} c_2 \xrightarrow{\text{spec}}*c'$ is preempted if there exists a path $c \xrightarrow{\text{spec}}*c'_1 \xrightarrow{f} c_1 \xrightarrow{\text{spec}}*c'$ such that $c_2 \xrightarrow{\text{spec}}*c_1$ (Figure 3.b).*

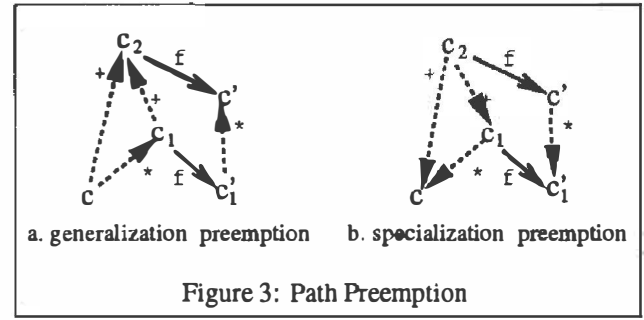

a. generalization preemption    b. specialization preemption

Figure 3: Path Preemption

Given a set of observations, an abductive reasoner seeks the best explanation of how the observed events might have been caused by certain culprit events. For example, to explain the sneezing and headache of Bob and those of Arnold after his contact with Bob, we may construct the scenario in Figure 4. This scenario states that Bob's flu caused his sneezing, headache and Arnold's flu, which in turn caused Arnold's sneezing and headache.

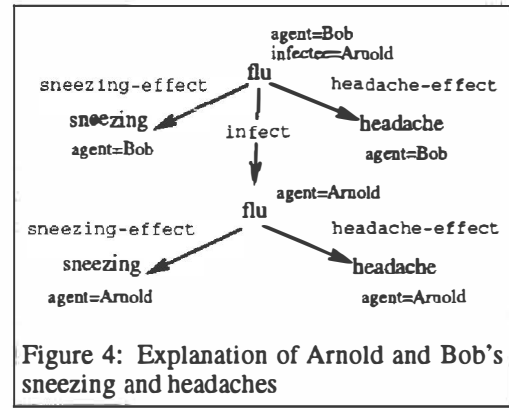

Figure 4: Explanation of Arnold and Bob's sneezing and headaches

The difference between a scenario and a subtree in a network is that a scenario may contain multiple instances of the same node in the network. Formally, a scenario is recursively defined with local trees:

**Definition 3.2 (Local Tree)** *A local tree is a one level tree whose nodes are event descriptions and whose links are labeled with distinct feature functions such that if there is a link labeled f from D to D' in the local tree, where $D(x) \equiv c(x) \wedge V(x)$ and $D'(x) \equiv c'(x) \wedge V'(x)$ are* *object descriptions, then there is a non-preempted path*
$c \xrightarrow{\text{lnk}} *c'' \xrightarrow{\text{f}} c'$ *in the event network.*

**Definition 3.3 (Scenario)** *A scenario is defined recursively as follows:*

1. *The local trees are also scenarios.*

2. *Suppose $\alpha$ is a scenario. If $D$ is a leaf node of $\alpha$ and $\beta$ is a local tree or a specialization link, whose root is also $D$, then the tree formed by joining the node $D$ in $\alpha$ and the root of $\beta$ is also a scenario.*

3. *All scenarios are either one of the above.*

The scenario in Figure 4 can be decomposed into two local trees. The first one consists of Bob's flu causing his sneezing, headache and Arnold's flu. The second one consists of Arnold's flu causing his sneezing and headache.

A scenario represents a formula that is the conjunction of the event descriptions in it. For example, the scenario in Figure 4 represent the following formula:

$\alpha(x) \equiv \mathtt{flu}(x) \wedge V(x) \wedge \mathtt{sneezing}(g_1(x)) \wedge V_1(g_1(x))$
$\wedge \mathtt{headache}(g_2(x)) \wedge V_1(g_2(x)) \wedge \mathtt{flu}(g_3(x))$
$\wedge V_2(g_3(x)) \wedge \mathtt{sneezing}(g_4(x)) \wedge V_2(g_4(x))$
$\wedge \mathtt{headache}(g_5(x)) \wedge V_2(g_5(x)$

where

$V(x) \equiv \mathtt{agent}(x) = \mathtt{Bob} \wedge \mathtt{infectee}(x) = \mathtt{Arnold}$
$\qquad \wedge \mathtt{time}(x) = 1$
$V_1(x) \equiv \mathtt{agent}(x) = \mathtt{Bob} \wedge \mathtt{time}(x) = 1$
$V_2(x) \equiv \mathtt{agent}(x) = \mathtt{Arnold} \wedge \mathtt{time}(x) = 2$
$g_1(x) \equiv \mathtt{sneezing}(x)$
$g_2(x) \equiv \mathtt{headache}(x)$
$g_3(x) \equiv \mathtt{infect}(x)$
$g_4(x) \equiv \mathtt{sneezing}(\mathtt{infect}(x))$
$g_5(x) \equiv \mathtt{headache}(\mathtt{infect}(x))$

**Definition 3.4 (Explanation)** *An explanation of a set of observations is a scenario of the event network that is consistent with the knowledge base and logically entails the observations.*

Let $\alpha(x)$ denote a scenario, the probability of the scenario is $[\alpha(x)]_x$. The problem of finding a causal explanation of the observations amounts to finding the most probable scenario that explains the observations.

Let $f$ be a feature that represents a causal tendency. When $f(x) \neq \bot$, it means that event $x$ caused another event $f(x)$. We call $f(x) \neq \bot$ a **causation event**. Assuming that given the direct cause of causation event, its probability is independent of indirect causes and other effects of the direct cause, we showed in [Lin, 1992] that the probability of a scenario is the product of the causation event involved in the scenario and the prior probability of the culprit (the root node of a scenario), which are available as category statistics in the knowledge base.

For example, if $\alpha(x)$ denotes the scenario in Figure 4, then

its probability is:

$$[\alpha(x)]_x = p_0 \times p_1 \times p_2 \times p_3 \times p_1 \times p_2$$

where

$p_0 = [\mathtt{flu}(x)]_x$
$p_1 = [\mathtt{sneezing}(\mathtt{sneeze\text{-}effect}(x))|\mathtt{flu}(x)]_x$
$p_2 = [\mathtt{headache}(\mathtt{headache\text{-}effect}(x))|\mathtt{flu}(x)]_x$
$p_3 = [\mathtt{flu}(\mathtt{infect}(x))|\mathtt{flu}(x)]_x$

# 4    INFERENCE ALGORITHM

Since we have used only a restricted subset of **Lp** to represent the observations, scenarios and domain knowledge, a specialized inference algorithm can be used to find the most probable explanation. Abductive inference is performed by a message passing algorithm operating over the network. Each of the nodes in the network is also a computing agent that communicates with neighbouring agents by passing messages across the links in the network. The messages represent partial explanations of the observations, which are explanations of a subset of observations. The partial explanations are combined with others at the nodes to construct explanations for larger subsets of observations. Dynamic programming techniques are employed so that the most probable scenarios can be found without enumerating all those possible. A detailed description of the algorithm can be found in [Lin, 1992].

# 5    PLAN RECOGNITION

Dynamic Bayesian network construction is most often used in plan recognition. We now use an example to demonstrate that we can represent domain knowledge in a static network and use it directly in plan recognition.

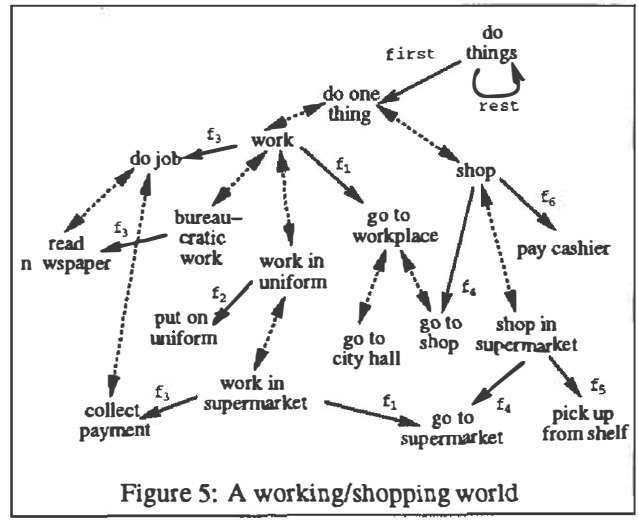

Figure 5: A working/shopping world

Figure 5 is shows a network representation of a fraction of working/shopping world. We briefly explain the knowledge encoded in the network:

- To work, one first goes to work place and then do whatever job he/she is suppose to do. Thus the feature

links from work to goto-workplace and do-job. For bureaucrats, doing their job means reading comics in newspapers.

- Some kinds of work require the workers to wear uniforms. Supermarket casher's work is one of them.

- A casher goes to supermarket to work and the job is to collect payment.

- To shop, one first goes to a shop, get the merchandise, and then pay the casher.

- Shopping in supermarket in a kind of shopping, where one goes to a supermarket and get the merchandise by picking them up from the shelves.

Once the network is created, we can use it to understand the stories about the domain. Consider the following example from [Ng and Mooney, 1990]:

"John went to the supermarket. He put on the uniform."

The two sentences are represented as two observations:

$\exists x.\text{go-to-supermarket}(x) \wedge \text{agent-name}(x) = \text{John}$
$\exists x.\text{put-on-uniform}(x) \wedge \text{agent-sex}(x) = \text{male}$

Two scenarios explaining the observations are shown in Figure 6. The scenario in (1) explains the observation by postulating that John is going to supermarket to work. The scenario (2), on the other hand, says that John went to the supermarket to shop and some man (not necessarily John) put on uniform to work. Assuming the probabilities:

$[\text{goto-supermarket}(f_1(x))|\text{work-in-supermarket}(x)]_x$
$[\text{put-on-uniform}(f_2(x))|\text{work-in-uniform}(x)]_x$
$[\text{goto-supermarket}(f_4(x))|\text{shop-in-supermarket}(x)]_x$
$[\text{do-one-thing}(first(x))|\text{do-things}(x)]_x$

are 1.0, the probability of scenario (1) is

$$[\text{work-in-supermarket}(x)]_x$$

whereas the probability of scenario (2) is

$[\text{do-things}(x)]_x \times$
$[\text{do-things}(rest(x))|\text{do-things}(x)]_x \times$
$[\text{shop-in-supermarket}(x)|\text{do-one-thing}(x)]_x \times$
$[\text{work-in-uniform}(x)|\text{do-one-thing}(x)]_x$

Therefore, the first scenario tends to be more probable than the second unless work-in-supermarket is an extremely unlikely event.

The event network in Figure 5 can also be used to understand many other stories in the same domain without having to construct new networks:

- From "Sue when to the shop, she picked up bottle of milk form the shelf and paid the cashier," we may infer that Sue bought a bottle of milk in a supermarket (Figure 7). Note that the story did not say the shop Sue went to is a supermarket. However, since Sue picked up the milk from the shelf herself, instead of asking an attendant to do it for her, we can infer that she was shopping in a supermarket.

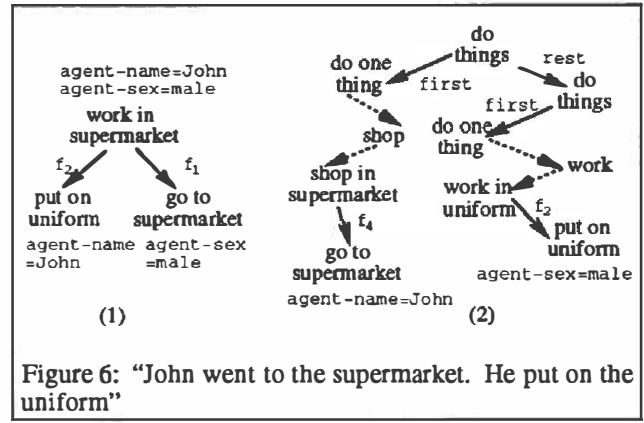

Figure 6: "John went to the supermarket. He put on the uniform"

- From "Bill went to the city hall and began to read newspaper," we may infer that Bill is a bureaucrat working in the city hall (See Figure 8). There is no direct explanation of going to city hall, the observation inherits explanation from going to work place.

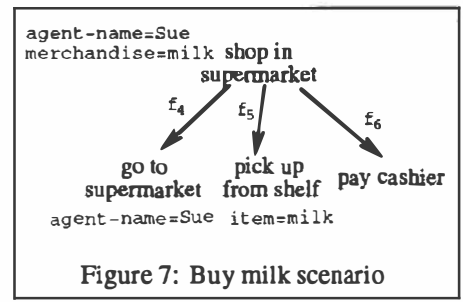

Figure 7: Buy milk scenario

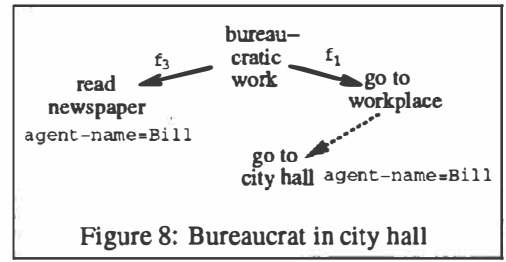

Figure 8: Bureaucrat in city hall

The ability to deal with networks with directed cycles also allows us to represent and recognize recursive plans. In [Lin, 1992], we showed that our scheme can be used for program recognition. We represent programs as hierarchies of recursive plans. Given a sequence of basic actions performed by a program, e.g., comparing and exchanging positions of two elements in an array, our plan recognition algorithm is able to infer which algorithm (plan) the program is using.

## 6  IMPRECISE OBSERVATIONS

Consider an example from auto-engine diagnosis: misfiring (mf) causes uneven sound (us). There are two possible causes of mf: bad spark (bs) and impure fuel (im). The former is an ignition system problem and tends to happen continually. On the other hand, the latter tends to be inter-mittent. The causal relationships between these events can be represented in the causal network in Figure 9.

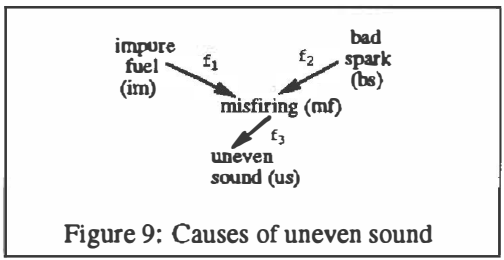

Figure 9: Causes of uneven sound

Since the occurrence attribute occ of bs is continual (cont), we have a local constraint at the node bs:

$$\forall x.\mathtt{bs}(x) \rightarrow \mathtt{occ}(x) = \mathtt{cont} \ldots\ldots\ldots(a)$$

On the other hand, since im is intermittent (int), we have:

$$\forall x.\mathtt{im}(x) \rightarrow \mathtt{occ}(x) = \mathtt{int} \ldots\ldots\ldots(b)$$

We also have percolation constraints:

$$\forall x, v.\mathtt{bs}(x) \wedge \mathtt{mf}(f_1(x)) \wedge \mathtt{occ}(f_1(x)) = v \rightarrow \mathtt{occ}(x) = v$$
$$\forall x, v.\mathtt{im}(x) \wedge \mathtt{mf}(f_2(x)) \wedge \mathtt{occ}(f_2(x)) = v \rightarrow \mathtt{occ}(x) = v$$
$$\forall x, v.\mathtt{mf}(x) \wedge \mathtt{us}(f_3(x)) \wedge \mathtt{occ}(f_3(x)) = v \rightarrow \mathtt{occ}(x) = v$$

Then if the observation is $\exists x.\mathtt{us}(x)$, then we have two explanations in Figure 10. Both bs and im are possible explanations.

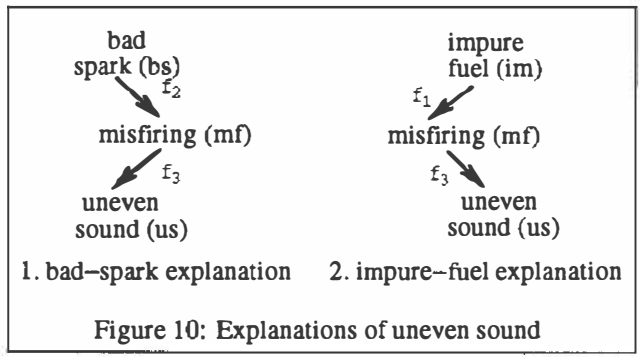

Figure 10: Explanations of uneven sound

If the observation is given in more detail:

$$\exists x.\mathtt{us}(x) \wedge \mathtt{occ}(x) = \mathtt{cont}$$

then the attribute value $\mathtt{occ} = \mathtt{cont}$ percolates from us to mf and then to bs and im (see Figure 11). In Figure 11.2, the attribute value $\mathtt{occ} = \mathtt{cont}$ at the node im violates the local constraint (b). Therefore the scenario in Figure 11.2 is inconsistent with the knowledge base and subsequently eliminated as a candidate explanation. On the other hand, the scenario in Figure 11.1 is consistent with the knowledge base and logically implies the observations. Therefore, it is an explanation of the observation.

# 7   SPECIFICITY OF EXPLANATION

In our formulation of abduction, specificity of explanation is treated as follows: Each individual explanation is maxi-mally specific with respect to the observations it explains, because explanations do not contain preempted path. Since the more general explanations are more probable, [1] the most general explanation that logically entails the observations is preferred.

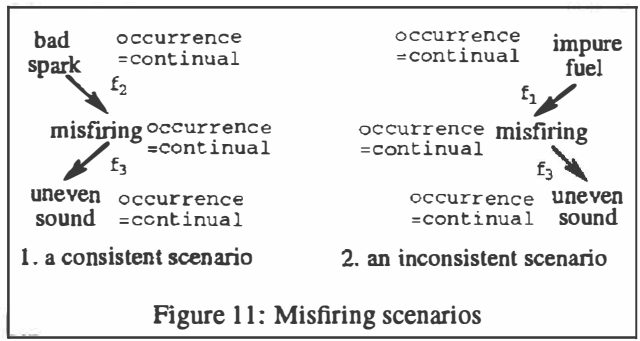

Figure 11: Misfiring scenarios

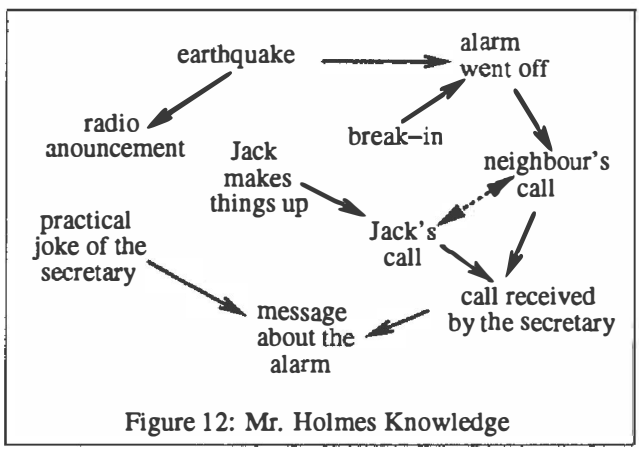

Figure 12: Mr. Holmes Knowledge

For example, suppose Mr. Holmes knowledge about phone message, house alarm, etc.  represented in Figure 12. The shaded nodes represent events based upon which Mr. Holmes makes his decisions. The root node of an explana-tion must be one of the shaded nodes. They are similar to End nodes in Kautz's plan hierarchy [Kautz, 1987].

Suppose Mr. Holmes's only observation is the phone mes-sage, then he has three explanations shown in Figure 13. The scenario shown in Figure 14.2 is deemed to be less likely than the one in Figure 13.2 because the probabil-ity of the former is the probability of the latter times the probability of a random neighbour of Mr. Holmes being Jack. Now suppose Mr. Holmes has an independent source confirming that Jack made a call, then the two explanations are those in Figure 14. The scenario in Figure 13.2 is no longer an explanation because it does not explain Jack's call. If Mr. Holmes further heard on radio that there was an earthquake, then his explanation became the scenario in Figure 15. Earthquake was not considered as explanations previously because its prior probability is much lower than the prior of break-ins. The scenario becomes preferable be-cause other scenarios do not explain radio announcement.

---

[1]By a more general explanation, we do not mean an explanation that explain a wider range of data, rather, we mean the explanation is stated in more general terms.

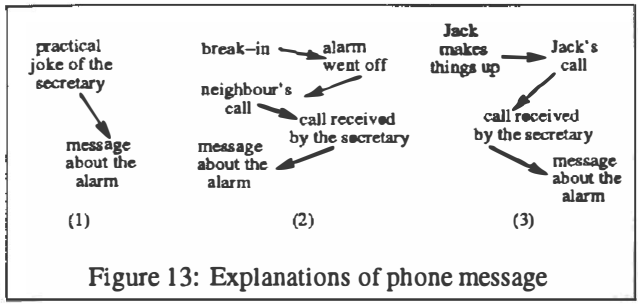

Figure 13: Explanations of phone message

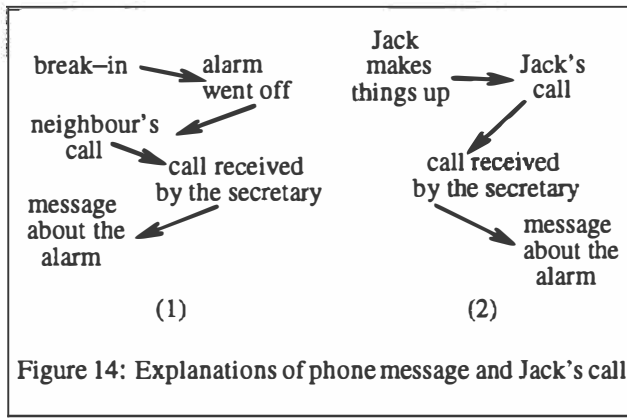

Figure 14: Explanations of phone message and Jack's call

# 8    CONCLUSION

We pointed out several drawbacks of Bayesian networks in abduction, due to its propositional nature and acyclic structure. We proposed a predicative interpretation of probabilistic network which is similar to Bayesian network in that the independence relationships between events are represented in a network structure but is different in that the nodes represent unary predicates instead of propositions. We have used several examples to show that our approach overcomes several difficulties encountered in Bayesian networks.

### Acknowledgement

This research was done at the University of Alberta. The author is indebted to Dr. Randy Goebel for his guidence.

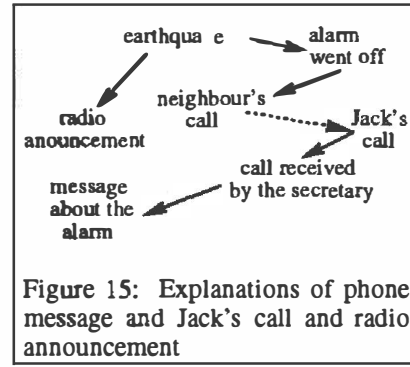

Figure 15: Explanations of phone message and Jack's call and radio announcement